%% file: root.tex
\documentclass[letterpaper, 10 pt, conference]{ieeeconf}  

\IEEEoverridecommandlockouts                              

\usepackage{amsmath} 
\usepackage{amssymb}  
\usepackage{multirow}
\usepackage{graphicx}
\usepackage{booktabs}
\usepackage{xcolor}
\usepackage{subcaption}

\title{MonoSLAM: Robust Monocular SLAM with Global Structure Optimization}

\author{Bingzheng Jiang$^{1,*}$, Jiayuan Wang$^{2,*}$, Han Ding$^{3}$, Lijun Zhu$^{1}$
\thanks{$^{*}$ Equal contribution.}%
\thanks{$^{1}$School of Artificial Intelligence and Automation, Huazhong University of Science and Technology, Wuhan, China, bzjiang@hust.edu.cn, ljzhu@hust.edu.cn.}%
\thanks{$^{2}$School of International Education, Wuhan University of Technology, Wuhan, China.}%
\thanks{$^{3}$School of Mechanical Science and Engineering, Huazhong University of Science and Technology, Wuhan, China.}%
}
\begin{document}

\maketitle
\thispagestyle{empty}
\pagestyle{empty}

\begin{abstract}

This paper presents a robust monocular visual SLAM system that simultaneously utilizes point, line, and vanishing point features for accurate camera pose estimation and mapping. To address the critical challenge of achieving reliable localization in low-texture environments, where traditional point-based systems often fail due to insufficient visual features, we introduce a novel approach leveraging Global Primitives structural information to improve the system's robustness and accuracy performance. Our key innovation lies in constructing vanishing points from line features and proposing a weighted fusion strategy to build Global Primitives in the world coordinate system. This strategy associates multiple frames with non-overlapping regions and formulates a multi-frame reprojection error optimization, significantly improving tracking accuracy in texture-scarce scenarios. Evaluations on various datasets show that our system outperforms state-of-the-art methods in trajectory precision, particularly in challenging environments. 

\end{abstract}


\input{Section/1_intro}

\input{Section/2_related}

\input{Section/3_method}

\bibliographystyle{IEEEtran} 
\bibliography{IEEEfull}

\end{document}

%% file: Section/1_intro.tex
\section{Introduction}
Tracking and reconstructing in unknown three-dimensional scenes based on visual inputs are fundamental tasks in robotics and computer vision. The performance of localization and mapping modules significantly impacts the service quality of robotic autonomous systems and augmented/virtual reality devices. However, these modules often suffer from pose drift in incremental camera tracking process.  
To address this problem, different strategies are proposed. On one hand, advanced sensors such as depth cameras, LiDAR, and IMUs can be used to provide more reliable information to enhance the performance of SLAM  and IMU-based  systems. On the other hand, algorithmic solutions like local bundle adjustment~\cite{mur2015orb,rosinol2020kimera}, sliding window optimization~\cite{qin2018vins,engel2017direct}, and loop closure techniques~\cite{labbe2019rtab,mur2015orb} help mitigate drift.  
The core optimization theories of such solutions are to explore leveraging \textbf{visual overlaps} to construct co-visibility factor graphs for optimization. But for monocular sensors that are affordable and widely utilized in devices, limited co-visible features can be utilized in tracking.   
Therefore, a key challenge remains in the community: how to capture and utilize more global information from monocular inputs to improve SLAM performance.


Point features have long been the cornerstone of most visual pose estimation systems, as evidenced by their widespread application in several leading methodologies \cite{mur2015orb,qin2018vins,rosinol2020kimera}. Despite the prevalence and successes of point features, they exhibit notable limitations, especially in environments with challenging scenarios, such as indoor spaces. In these contexts, the lack of distinct and rich point features often obstructs the real-time tracking capabilities of SLAM systems, leading to the need for alternative strategies. For example, the robustness of point-only factor graph optimization~\cite{li2023open} faces degeneration that can be partially enhanced by leveraging line~\cite{lu2015visual,zuo2017robust} and plane~\cite{zhou2021lidar} landmarks into tracking and optimization modules. 
Plane detection generally requires the use of depth maps \cite{salas2014dense} or convolutional neural networks \cite{paigwar2020gndnet}. In contrast, lines can be easily extracted from RGB images, offering a more versatile and resource-efficient alternative for incorporating information into visual odometry systems. And the most widely used line parameterization in line-related SLAM systems is the \textit{Orthonormal} algorithm~\cite{bartoli2005structure}, an elegant strategy based on the \textit{Lie Group} and \textit{Lie Algebra}.

\begin{figure}
    \centering
    \includegraphics[width=\linewidth]{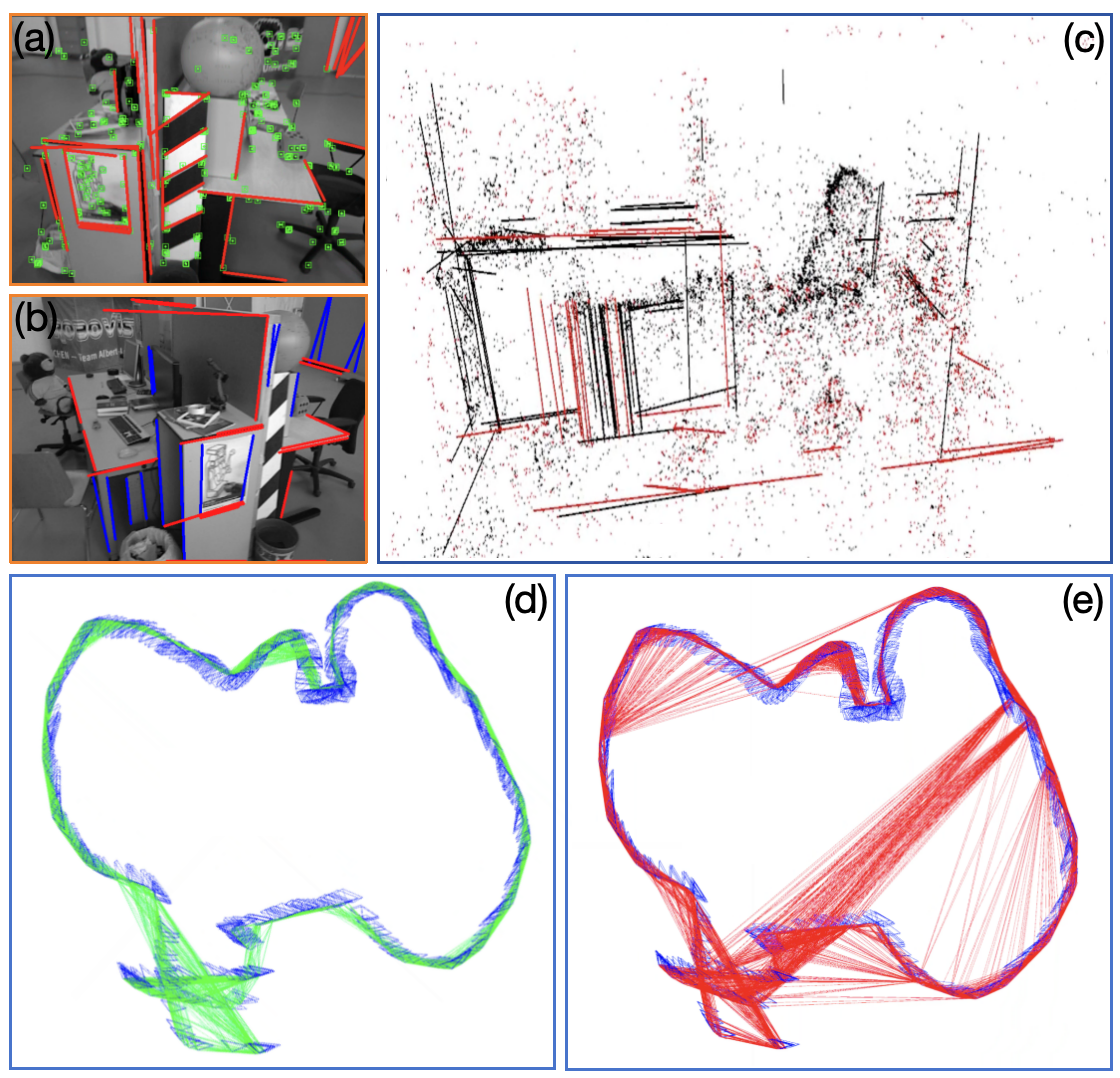}
    \caption{Example results of the proposed method. (a) Point and line features extracted from a single RGB image; (b) Segmented lines that are associated with Global Primitives; (c) 3D sparse map established by point and line landmarks; (d) and (e) Connected frames linked via green and red connections, respectively, to build up the proposed covisibility graph and Global Primitive association graph.}
    \label{fig:teaser}
\end{figure}

Traditionally, a single line segment contributes a re-projection factor to the optimization module, whereas a collection of line segments provides broader structural regularities. Specifically, a cluster of parallel line segments on a 2D image plane converges at a vanishing point, which can serve as a constraint for factor graph optimization.  
By assuming an Atlanta/Manhattan World environment, 3D line landmarks can be leveraged to establish perpendicular and orthogonal pairs, aiding in orientation estimation. A group of orthogonal vanishing direction vectors constructs a Manhattan World structure~\cite{straub2017manhattan}, where the assumption supports drift-free rotation estimation strategies~\cite{li2020structure,kim2018low} in visual odometry methods, but it is difficult to optimize the structure as one primitive in factor graph optimization modules. 
Additionally, line landmarks can provide co-planarity constraints when 3D planes are detected from sparse point landmarks extracted from monocular images.  
However, re-projection constraints derived from 3D lines or vanishing points are primarily effective for adjusting local regions, as a 3D line landmark is typically visible in only a limited number of frames. While Atlanta/Manhattan World assumptions provide global cues across frames, they face significant challenges in complex, unstructured environments.


To solve this issue, the method proposes a novel SLAM architecture for monocular inputs, which first explores global and flexible structural primitives from sequential images, and then build new factor graph to optimize camera poses and global primitives. Compared to traditional factors, the proposed constraints can be constructed between images without visual overlaps.
The contributions can be summarized as:

\begin{itemize}
    \item A high-precision real-time monocular SLAM framework without environmental structural constraints, extracting point features, line features, and vanishing point features from images.
    \item A multi-frame non-overlapping region image association strategy based on global primitives.
    \item An effective factor graph optimization combining global primitives for high-accuracy pose estimation.
\end{itemize}

%% file: Section/2_related.tex
\section{Related Work} \label{sec:related_work}

\subsection{Landmark Representation}
When we discuss landmark reconstruction and optimization tasks, two stages of representation are considered in the process, where the first is used in triangulation from 2D measurements while the goal of the second is finding a minimal parametrization for the iterative refinement process. When the degrees of freedom used in reconstruction satisfy the minimal parametrization requirements, those two stages can be merged in one step.
For point landmarks, the Euclidean \textit{XYZ} form representation is popularly used in the reconstruction and optimization modules of SLAM approaches~\cite{mur2015orb,qin2018vins}. To provide a wide range of depths to point features, inverse depth parametrization~\cite{civera2008inverse} is proposed relative to the camera poses from which they were first viewed and related measurements. 

When we use the Euclidean \textit{XYZ} method to parametrize endpoints of a finite line in 3D space, the representation can be used to provide re-projection residuals of lines for camera pose optimization, but it has over-parameterization problems in landmark optimization. Similar to the \textit{Euclidean}, a line at infinity can be represented as the \textit{Pl\"ucker} coordinate via two three-dimensional vectors containing the direction of the line and normal based on the line and the camera coordinate frame. Furthermore, the dual quaternion approach~\cite{kottas2013efficient}  represents a line in 3D space using dual quaternions, which provides a concise way to represent rotations and translations in 3D space and can be used to encode the position and orientation of lines.

\subsection{Structural Regularities}
Indoor environments, typified by structures formed by lines and planes, are integral to constraining factor graph optimization processes. In this context, the Manhattan World (MW) and Atlanta World (AW) assumptions are pivotal in simplifying pose estimation tasks. These assumptions enable a reduction in non-linearity by computing orientations using structured environments. 
For instance, Structure-SLAM ~\cite{li2020structure} leverages surface normals predicted from monocular RGB images for orientation estimation, while Linear-SLAM ~\cite{joo2021linear} relies on extracting planes from depth maps. 
Given a known rotation, the task of estimating the translation becomes linear.
The MW is defined by an orthonormal set of landmarks, whereas the AW is characterized by a vertical direction and multiple horizontal directions perpendicular to it. The multi-Manhattan World model~\cite{yunus2021manhattanslam} introduces a more flexible approach, postulating that only certain local regions, such as wall corners, need to adhere to orthogonal relationships, versus the entire scene conforming to a single MW paradigm. However, these methods tend to overlook the optimization of these structural elements within the factor optimization framework. E-Graph~\cite{li2022graph} provides the extensibity graphs, but these graphs are only used in initial pose estimation instead of factor graph optimization.

Departing from scene-specific structural models, recent methodologies have pivoted towards integrating structural landmark constraints, particularly lines and planes, directly into optimization loss functions. 
~\cite{zhang2015building} were pioneers in this regard, introducing parametric 3D straight lines as structural landmarks in SLAM systems. This approach significantly improves the traditional SLAM methodology, which relied primarily on points, by offering a more detailed environmental representation.
~\cite{lu2015visual} expanded this concept thought employing a diverse array of structural features, including points, line segments, extended lines, planes, and vanishing points. These features, along with their intrinsic geometric constraints, were integrated as high-level landmarks within a multilayer feature graph (MFG), which offers a more comprehensive representation of the environment and improves pose estimation.
Kimera-VIO ~\cite{rosinol2020kimera} represents another advance, utilizing 3D mesh information to detect coplanarity, thus infusing structural regularities into the factor graph. This method illustrates that using structural information to improve the accuracy of 3D landmarks can simultaneously improve the accuracy of pose estimation.
Building on the Kimera-VIO mesh methodology, PLP-VIO ~\cite{li2020leveraging} introduces a novel approach by using line segments to construct meshes with edge information. This technique, which contrasts with Kimera-VIO's point-only mesh construction, incorporates coplanar constraint structures of straight lines and point lines, enriching spatial understanding.
Furthermore, CoP ~\cite{li2020co} innovates by creating a new parameterization for constraint groups by representing points and lines based on plane parameters. When these constraints are incorporated into loss functions, they ensure the preservation of essential geometric relationships throughout the optimization stage, demonstrating the growing sophistication and effectiveness of current SLAM methodologies.


%% file: Section/3_method.tex
\begin{figure*}
\vspace{.3cm}
    \centering
    \includegraphics[width=\linewidth]{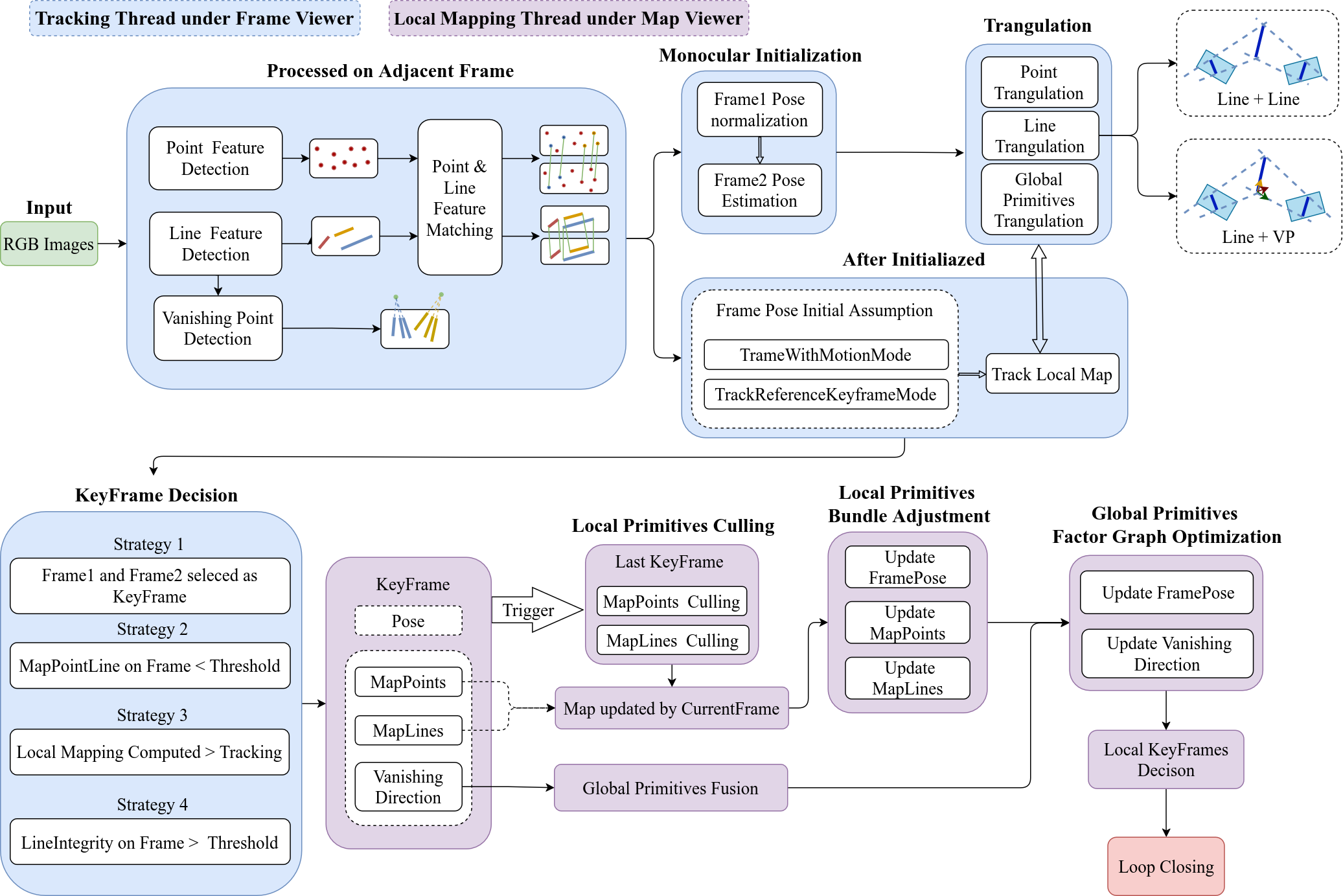}
    \caption{Architecture of the Proposed System. The MonoSLAM framework comprises a robust front-end and a multi-level back-end. In the front-end, point, line, and vanishing point features are extracted from the RGB image to establish a rich representation of the scene. In the back-end, the framework first leverages the scene structure, represented by points and lines, to estimate the camera pose and update the map. Subsequently, global primitive constraints are integrated to further refine the camera pose estimation and enhance the map accuracy. This dual-stage optimization approach ensures precise and reliable performance in complex environments.}
    \label{fig:archi}
\end{figure*}

\section{system overview}

As shown in Figure~\ref{fig:archi}, the system contains front-end and back-end modules. In the front-end, we introduce the detection methods for points, lines, and vanishing points. And then, in our back-end, the first part is a new association strategy where on the one hand 
points and lines are utilized to generate 3D landmarks that are \textbf{Local Primitives (LP)} collected in the map, while on the other hand, we provide a novel strategy to generate \textbf{Global Primitives (GP)} structure information during the association process.  
After detecting local and global primitives, the second part of the back-end builds a robust factor graph optimization algorithm for accurate camera pose estimation.

\subsection{Feature extraction and Vanishing point detection}
In feature detection step, ORB~\cite{mur2015orb} is employed for the extraction of feature points. Given the challenge posed by the scarcity of high-quality feature points in low-texture environments for pose estimation, we further introduced the LSD~\cite{von2008lsd} and LBD~\cite{zhang2013efficient}  algorithms to extract and describe line features within the images. Additionally, to enhance the robustness of the algorithm, the JLinkage~\cite{toldo2008robust} algorithm was utilized to extract vanishing points and their associated line features from the images. Specifically, the process begins with the random sampling of all extracted line features in the image to generate vanishing point hypotheses. Subsequently, by comparing and merging these hypotheses, the final vanishing points and their corresponding sets of line features are computed.

\subsection{Local Primitives }
The 3D point $\mathbf{P}_w$, in the world coordinates, can be represented as $\mathbf{P}_w=[ \mathbf{x} \; \mathbf{y} \;\mathbf{z}]^T$, where $\mathbf{x}$, $\mathbf{y}$ and $\mathbf{z}$ are coordinate values on each coordinate axis.
The 3D line can be represented based on the Pl\"ucker method, $\mathcal{L}=[ \mathbf{n}^T \; \mathbf{d}^T ]^T$, where $\mathbf{n}$ and $\mathbf{d}$ are normal and direction components of $\mathcal{L}$, respectively.
\subsection{Global Primitives}
We extract the vanishing point from the image and obtain the initial vanishing direction $\mathbf{V}_D^{w}$ in the world coordinate through the transformation matrix and camera intrinsic parameters. Then, we obtain the Global Primitives by performing a weighted fusion of $\mathbf{V}_D^{w}$ in the world coordinate system.
\subsection{Robust Factor Graph Optimization}
We constructed a robust factor graph optimization based on the reprojection errors of both Local Primitives and Global Primitives. Particularly during the optimization process for Global Primitives, the joint optimization of multiple non-overlapping frames can effectively improve the accuracy of pose estimation.

\section{Methodology}

\subsection{Global Structure Cues}


\paragraph{Optical Flow-based Line Feature Fusion}
To enhance the quality of the extracted line features, we employ optical flow-based line prediction to consistently track line features~\cite{lim2021avoiding}. First, we remove excessively short line features within the image when the length of the segments less than $\tau_s$ to ensure the robustness of line prediction. After detecting 2D lines $\mathbf{l}_j$ in frame $\textit{I}_{t-1}$, we predict the endpoints of the line feature in frame $\textit{I}_{t}$ by leveraging the optical flow map. By using the geometric characteristics of line features and their LBD descriptors, we match the predicted line features with those directly extracted from the frame $I_t$. In cases of successful matches, we compare the matching scores between the extracted line features and their corresponding optical flow-predicted counterparts, selecting the one with the higher score as the representative line feature in the frame $I_t$. To address the issue of fragmented line features where a single continuous line is segmented into multiple shorter lines during extraction, we merge the line features from time $t-1$ with their updated counterparts at time $t$ obtained through line prediction. This merging process aims to produce more stable and information-rich line features at time $t$. When merged line segments extend beyond the image boundaries, we choose to retain these out-of-bounds line features to enrich the set of line features for subsequent mapping modules.



\paragraph{Vanishing  direction}
We extract the vanishing point $\mathbf{v}$ of 2D straight lines from the image with JLinkage in the pixel coordinate. Subsequently, we utilize the transformation matrix and the camera intrinsic parameters $\mathbf{K}$ to obtain the vanishing direction $\mathbf{V}_D^{w}$ in the world coordinate of the image via the following formula:
\begin{equation}
    \mathbf{V}_D^{w} = \mathbf{R}_{wc} \pi(\mathbf{v}, \mathbf{K})
\end{equation}
where $\pi()$ transforms the vanishing point to vanishing direction in the camera coordinates.

Based on this, we can associate multiple keyframes in the world coordinate, and perform weighted fusion on the vanishing  directions in the world coordinate that are parallel to each other.

For each frame, the same number of line features are consistently extracted, and this count is denoted as $\rm{N_l}$. Given that $\mathbf{V}^w_{Di}$ and $\mathbf{V}^w_{Dj}$ are mutually parallel, $N_{v^w_Di}$ and $N_{v^w_Dj}$ denotes the size of a set of 2D lines associated with the global vanishing direction, $\mathbf{V}^{w^{\prime}}_{Di}$ and $\mathbf{V}^{w^{\prime}}_{Dj}$ represent the corrected values, respectively:
\begin{equation}
    \mathbf{V}^{w^{\prime}}_{Di} = \mathbf{V}^{w^{\prime}}_{Dj} = \frac{N_{v^w_Di}}{\rm{N_l}} \mathbf{V}^w_{Di} + \frac{N_{v^w_Dj}}{\rm{N_l}} \mathbf{V}^w_{Dj}
\end{equation}

\subsection{Local Mapping}
\paragraph{Keyframe Insertion}
In addition to conventional keyframe insertion strategies~\cite{campos2021orb}, we introduce a novel insertion criterion. Specifically, if optical flow-based prediction demonstrates the ability to consistently track a set of line features over an extended period, or if there is a substantial increase in the number of simultaneously tracked line features, the current frame is designated as a keyframe, triggering the execution of mapping operations. This mechanism effectively exploits the temporal continuity of line features, enhancing both the robustness of the system and the completeness of the reconstructed map.

\paragraph{Global Structure Association}

To ensure the quality of the constructed mapLines, we perform a filtering verification on the mapLines prior to mapping the line features. First, we use 
Reprojection Error Verification step to remove outliers. To be specific, the reprojection quality of feature lines is evaluated by comparing the angular error and the perpendicular distance error of the endpoints between the original pixel positions and the re-projected positions.
This is employed to determine whether the reprojection performance of the feature lines meets the required thresholds. Let $p^m_{\text{ori}}$ denote the midpoint of the original feature line in the image plane, and ${p}^m_{\text{proj}}$ represent the midpoint of the projected feature line in the image plane. The angular error $\theta _e$ is computed as the norm of the difference between these two midpoints in both the current frame and the reference frame:
\begin{equation}
     \theta_e = \Vert p^m_{\text{ori}} - p^m_{\text{proj}} \Vert
\end{equation}
where maplines satisfying  $\theta _e > \theta_{\text{thre}}$ are excluded from mapping, $\theta_{\text{thre}}$ denotes reprojection angular threshold. Let $d_s$ denotes the distance from the starting point of the original feature line to the projected feature line. Let $d_e$ denotes the distance from the ending point of the original feature line to the projected feature line. Let $d$ denotes as: 
\begin{equation}
     d = \max(d_s,d_e)
\end{equation}
Let $d_{\text{thre}}$ denotes the line distance threshold. If the perpendicular distance error is $d > d_{\text{thre}}$, the mapline is excluded from mapping.

After checking the matching relationship, Sensitivity Error Verification is further used to improve the robustness of association. Assuming, $v_{{\text{ori}}}$ denote the projected direction vector of the original 3D line, and $v_{\text{proj}}$ represents the projected direction vector calculated from the projection midpoint $p^m_{\text{proj}}$. The cosine of the angle $\alpha$ between these vectors is computed:
\begin{equation}
\cos\alpha  = \Vert v_{{\text{ori}}} \cdot v_{\text{proj}} \Vert
\end{equation}
where we convert  $\alpha$ to degrees. $\alpha_{\text{thre}}$ denotes the sensitivity threshold. If $ 90^{\circ} - \alpha > \alpha_{\text{thre}}$, the mapline is excluded from mapping.

Finally, we propose the Overlap Error Verification module that considers overlap quality of the feature line is evaluated by computing the degree of overlap between the projected position and the original pixel position. This is employed to determine whether the overlap condition of the feature line meets the required threshold. The projected length and projected direction vector of the original feature line are calculated as $l_{\text{ori}}$ and $v_{\text{{ori}}}$ respectively, followed by the computation of the projection ratio $r_1^{\prime}$ and $r_2^{\prime}$:
\begin{equation}
     r_1^{\prime}=\min(r_1,r_2), \quad 
     r_2^{\prime}=\max(r_1,r_2)
\end{equation}
where 
\begin{equation}
     r_{1} = \frac{(p^s_{\text{proj}} - p^s_{\text{ori}}) \cdot v_{\text{{ori}}}}{l_{\text{ori}}}
\end{equation}
\begin{equation}
     r_{2} = \frac{(p^e_{\text{proj}} - p^s_{\text{ori}}) \cdot v_{\text{{ori}}}}{l_{\text{ori}}}
\end{equation}
From (6), we can calculate $r$ as
\begin{equation}
     r = \min(r_2^\prime,1)-\max(r_1^\prime,0) 
\end{equation}
where $r$ denotes the overlap degree error, $p^s_{\text{ori}}$ and $p^e_{\text{ori}}$ denote the endpoints of the original feature line, and $p^s_{\text{proj}}$ and $p^e_{\text{proj}}$ represent the endpoints of the projected feature line. Let $r_{\text{thre}}$ denotes overlap degree threshold. If $r < r_{\text{thre}}$, the mapline is excluded from mapping.

\subsection{Factor Graph Optimization}

In this paper, we introduce a novel factor that leverages global primitives to improve the robustness of structure-based optimization. And we also use traditional factors in the optimization:

\paragraph{Point Reprojection Error}  
To ensure accurate alignment between 3D points and their corresponding image observations, we minimize the reprojection error $e_p$:  

\begin{equation}
    e_{p} = ||d_p(\Pi(\mathbf{P}_w, \mathbf{T}_{cw}), \mathbf{p})||
\end{equation}  
where \( \mathbf{P}_w \) denotes the 3D point in the world coordinate frame, \( \mathbf{T}_{cw} \) represents the camera pose, \( \Pi(\cdot) \) is the camera projection function, and \( \mathbf{p} \) is the observed 2D feature.  

\paragraph{Line Reprojection Error}  
For structural constraints, we extend the reprojection error $e_{l}$ to 3D line features represented in Plücker coordinates \( \mathcal{L}= [n \; d] \):  

\begin{equation}
    e_{l} = ||d_l(\Pi(\mathcal{L}_w, \mathbf{T}_{cw}), \mathbf{p}_{\text{proj}})||
\end{equation}  
where \( \mathcal{L}_w \) denotes the 3D line parameterized in the world frame, and \( \mathbf{p}_{\text{proj}} \) denotes its projected endpoints in the image.  

\paragraph{Global Structure Constraint}  
To further enforce global structural consistency, we introduce a constraint based on the global vanishing direction \( \mathbf{V}_D^{w} \) and its alignment with 2D line segments. The transformation follows:  

\begin{equation}
    \mathbf{v}_d = \mathbf{R}_{cw} \pi^{-1}(\mathbf{v}, \mathbf{K}) \\
\end{equation}

\begin{equation}
    e_{\mathbf{v}_d} = d_l( l{\in \mathcal{S}_{V_D^W}}, \mathbf{v}_d ) 
\end{equation}  
where \( \mathbf{R}_{c_w} \) transforms the vanishing direction from the world frame to the camera frame, \( K \) represents the intrinsic camera matrix, \( \mathbf{v}_d \) is the projected vanishing direction in the image space and  $\pi^{-1}()$ transforms the vanishing direction to vanishing point in the pixel coordinates. The error term \( e_{\mathbf{v}_d} \) ensures that the detected 2D line segments \( l \) belonging to the set \( \mathcal{S}_{V_D^W} \) are aligned with the global vanishing direction. $\mathcal{S}_{V_D^W}$ denotes the set of 2D lines associated with the global vanishing direction.

By integrating these geometric constraints, our approach enhances the accuracy and structural consistency of reconstructed scenes, particularly in environments with strong man-made structures.  
\textbf{Optimization with the Proposed Factor}  
In this paper, we propose a novel factor based on the global primitive:  

\begin{equation}
\mathbf{V}_D^{w} = \begin{bmatrix} n_x \\ n_y \\ n_z \end{bmatrix}
\end{equation}  
where $n_x$, $n_y$, and $n_z$ represent the components of $\mathbf{V}_D^{w}$ along the three coordinate axes in the world coordinate system, respectively.




\textbf{Tangent Space Optimization.} To ensure smooth optimization, we reparameterize the vanishing direction in the tangent space as follows:  
\begin{equation}
\delta\mathbf{d}=w_{1}\mathbf{b}_{1}+w_{2}\mathbf{b}_{2}
\end{equation}
where $\mathbf{b_1}$ and $\mathbf{b_2}$ are two orthogonal basis spanning the tangent space. $w_{1}$ and $w_{2}$ are corresponding displacements towards $\mathbf{b_1}$ and $\mathbf{b_2}$, respectively. The estimated direction \( \hat{d} \) is then expressed as a linear combination of the basis vectors in the tangent space:  

\begin{equation}
   \hat{d} = \delta\mathbf{d} + \mathbf{V}_D^{w}
\end{equation}  
After verifying these relationships, we define the structural consistency error $\mathbf{e}_{\text{str}}$:  
\begin{equation}
  \mathbf{e}_{\text{str}} = || \mathbf{L}_{D}^{w} \cdot  \mathbf{V}_D^{w} -1 ||
\end{equation}  
where $\mathbf{L}_{D}^{w}$ denotes the direction
vector of the original 3D line.

\paragraph{Nonlinear Least Squares Formulation}  
Finally, we formulate the overall optimization as a nonlinear least squares problem that incorporates point, line, and vanishing direction constraints by defining $E_\mu =\mathbf{e}_{\mathbf{\mu }}^\top \Sigma^{-1}_{\mathbf{\mu}} \mathbf{e}_{\mathbf{\mu }}$ where $\mu =\{p,l,v_d\}$. \( \Sigma_{\mathbf{p}}, \Sigma_{\mathbf{l}}, \) and \( \Sigma_{\mathbf{v}_d} \) represent the covariance matrices for the respective error terms, ensuring a balanced optimization process. The cost function to minimize can be written as follows:
\begin{equation}
    \mathcal{E} = \sum_pE_p + \sum_l E_l  +\sum_{v_d} E_{v_d}
\end{equation}
By integrating both LP and GP, this formulation enhances geometric consistency, leading to improved feature alignment and reconstruction accuracy.

\section{Experiments}

\subsection{Implementation Details}
To evaluate the proposed system, public datasets are used in this section to validate state-of-the-art methods and ours. All evaluations are conducted on a computer laptop equipped with an Intel Ultra9-285K CPU, ensuring consistent and reproducible results across all experiments.

\subsection{Baselines, Metrics, and Datasets}
We evaluate the mapping accuracy of our system by comparing it with state-of-the-art monocular SLAM systems. To validate the efficiency of our proposed line segments and vanishing points processing pipeline, we select structured image sequences from the \textbf{ICL-NUIM dataset}~\cite{handa2014benchmark}, which provides low-contrast and low-texture synthetic indoor sequences that are particularly challenging for monocular SLAM. The \textbf{root mean square error (RMSE)} is used as the primary metric, computed using the \textbf{evo toolkit}. The sequences \textbf{lr} and \textbf{of} represent the living room and office room scenarios in the ICL-NUIM dataset, respectively.  

We compare against six state-of-the-art systems on the ICL-NUIM dataset, which are GeoNet~\cite{qi2018geonet},  LPVO~\cite{kim2018low},  CNN-SLAM~\cite{tateno2017cnn}, LSD-SLAM~\cite{engel2014lsd}, Structure-SLAM~\cite{li2020structure} and ORB-SLAM3~\cite{campos2021orb}. GeoNet leverages geometric and photometric consistency to enhance pose estimation accuracy, particularly in dynamic environments, while CNN-SLAM~\cite{tateno2017cnn} and Structure-SLAM~\cite{li2020structure} integrate neural-based depth prediction and normal maps into the tracking modules, respectively. LPVO~\cite{kim2018low} focuses on efficient and real-time performance by optimizing parallel computation on modern hardware, and LSD-SLAM~\cite{engel2014lsd} employs direct methods for real-time dense mapping, eliminating the need for feature extraction. ORB-SLAM3~\cite{campos2021orb} is a feature-based SLAM system supporting monocular, stereo, and RGB-D cameras, with loop closure, relocalization, and map reuse capabilities.  

Additionally, we evaluate our method on the \textbf{EuRoC dataset}~\cite{burri2016euroc}, a widely used benchmark for visual SLAM. For this comparison, we focus on systems capable of utilizing multiple types of features, including point, line, and vanishing point features. The selected baselines tested on this dataset are PL-SLAM~\cite{gomez2019pl}, UV-SLAM~\cite{lim2022uv}, Struct-VIO~\cite{zou2019structvio}, PLF-VINS~\cite{lee2021plf}, Structure-PLP-SLAM~\cite{shu2023structure}, and AirVIO~\cite{xuairvo}. 
The first two systems that leverage line features using the LBD descriptor, and Struct-VIO tracks line features by sampling points along them. Structure-PLP-SLAM combines points, lines, and planes for enhanced robustness. PLF-VINS and AirVIO integrate visual features within a tightly coupled visual-inertial framework.




\begin{figure}
    \centering
    \begin{subfigure}{0.3\linewidth} 
        \includegraphics[width=\linewidth]{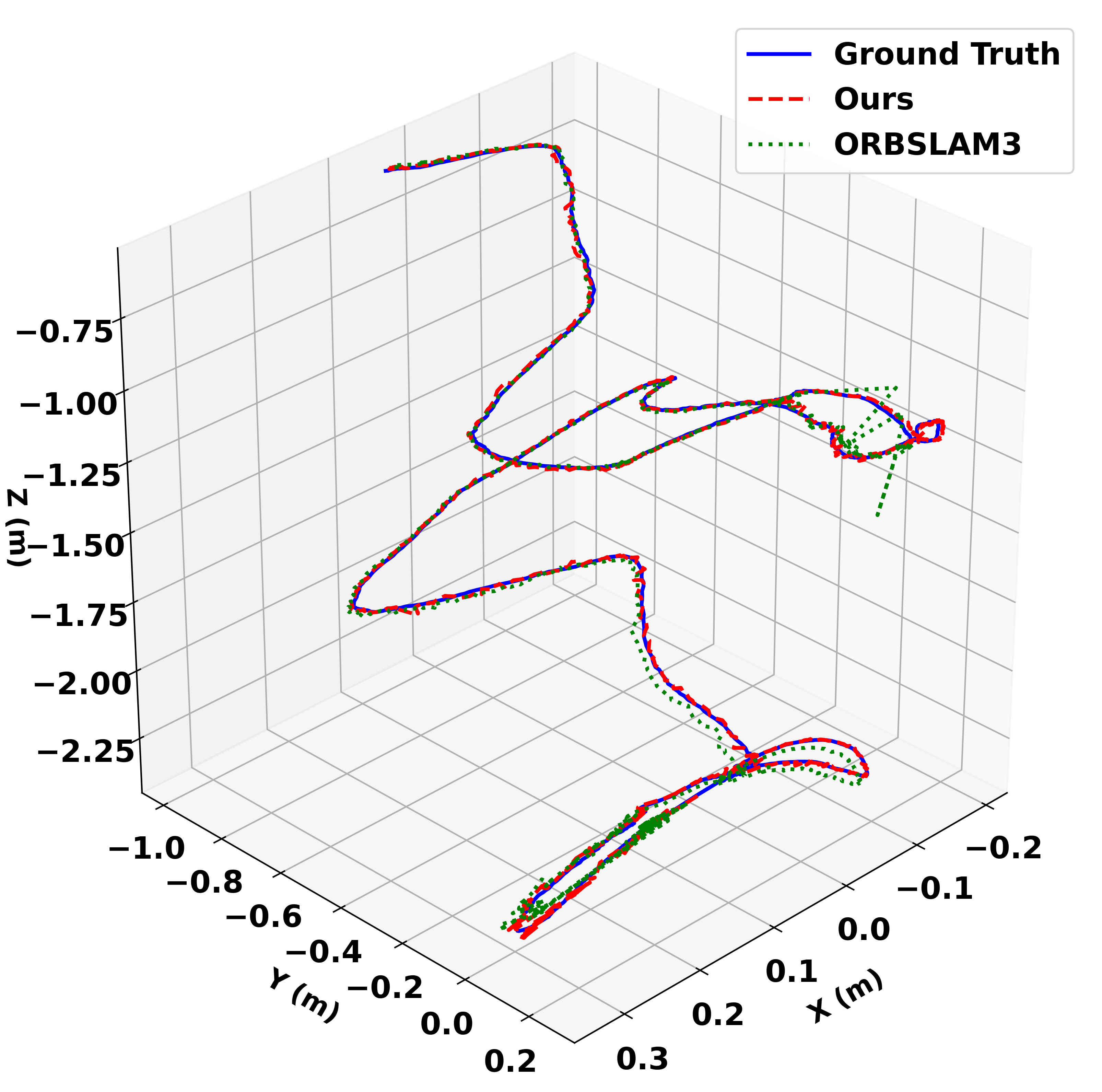}
        \caption{lr0}
        \label{fig:subfig1}
    \end{subfigure}
    \hfill
    \begin{subfigure}{0.3\linewidth}
        \includegraphics[width=\linewidth]{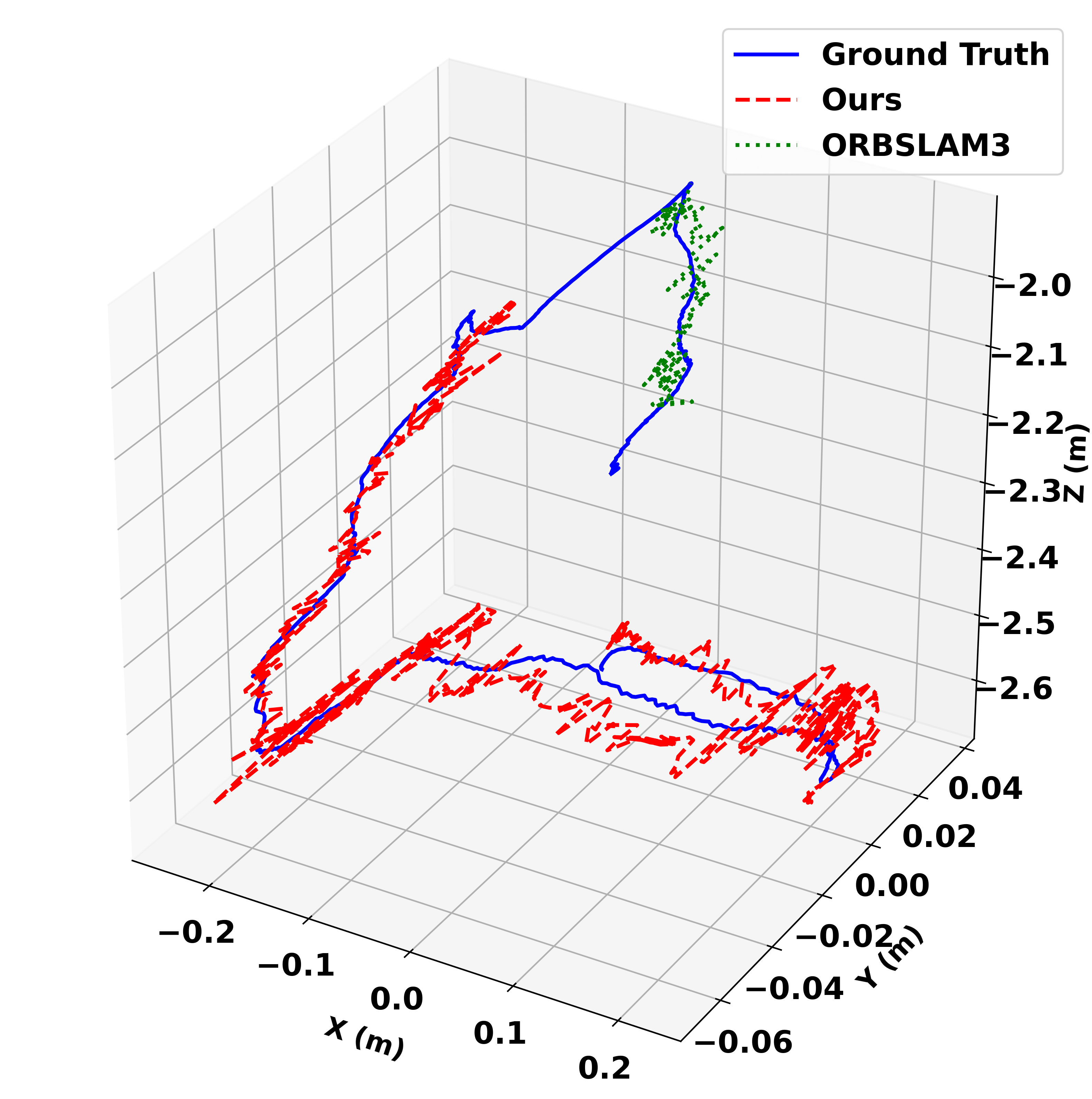}
        \caption{lr1}
        \label{fig:subfig2}
    \end{subfigure}
    \hfill
    \begin{subfigure}{0.3\linewidth}
        \includegraphics[width=\linewidth]{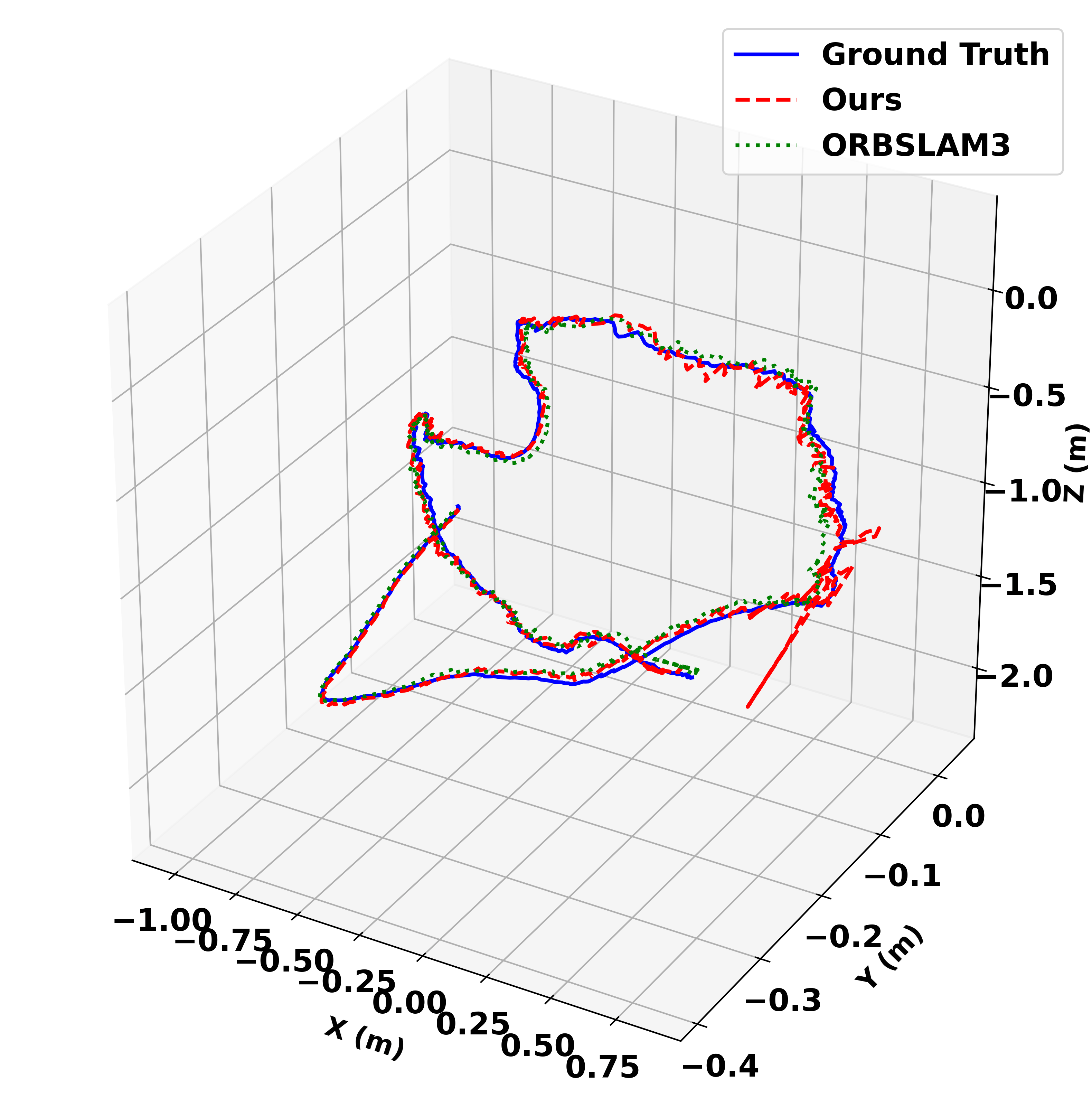}
        \caption{lr2}
        \label{fig:subfig3}
    \end{subfigure}

    \begin{subfigure}{0.3\linewidth} 
        \includegraphics[width=\linewidth]{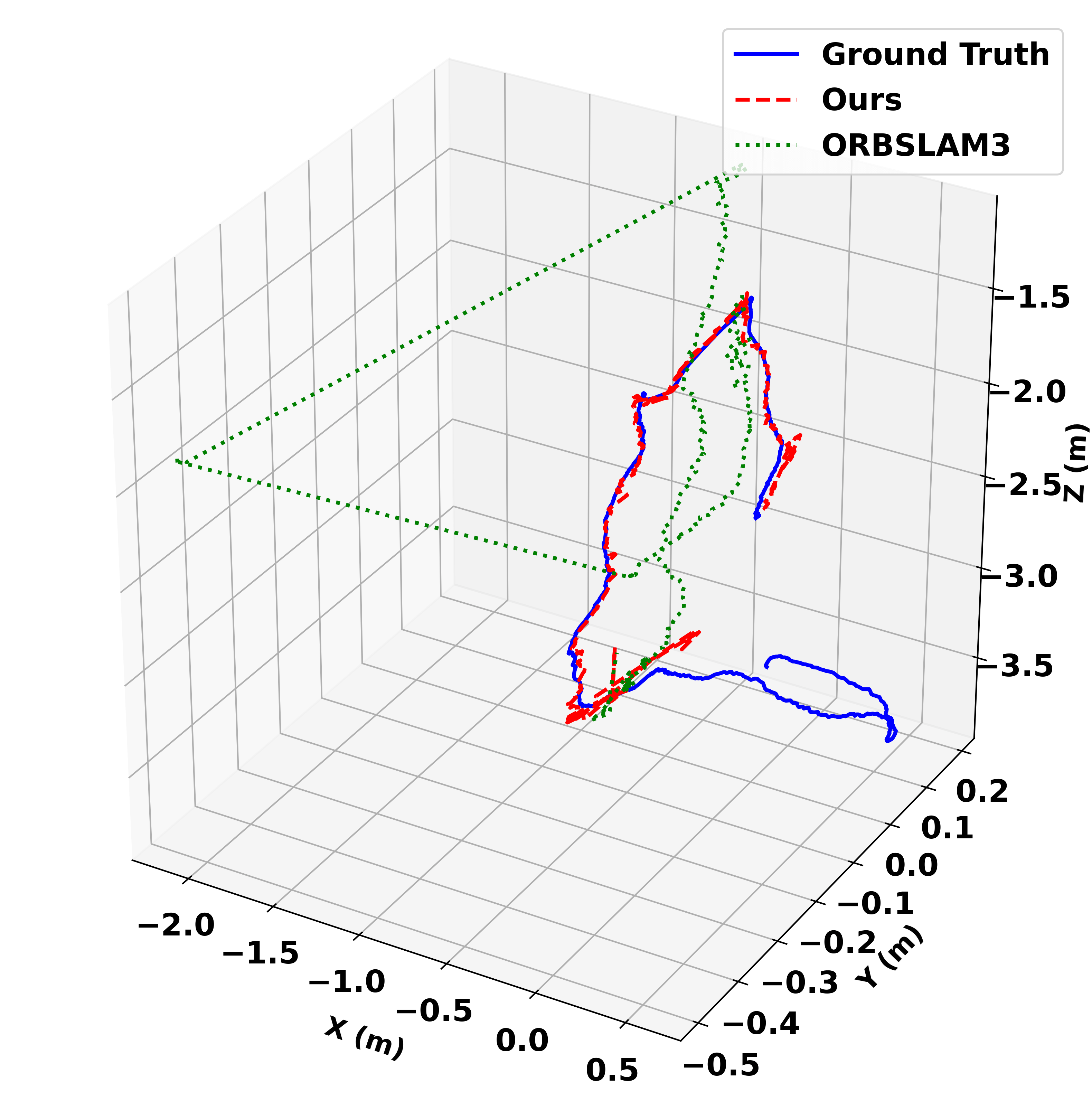}
        \caption{of1}
        \label{fig:subfig4}
    \end{subfigure}
    \hfill
    \begin{subfigure}{0.3\linewidth}
        \includegraphics[width=\linewidth]{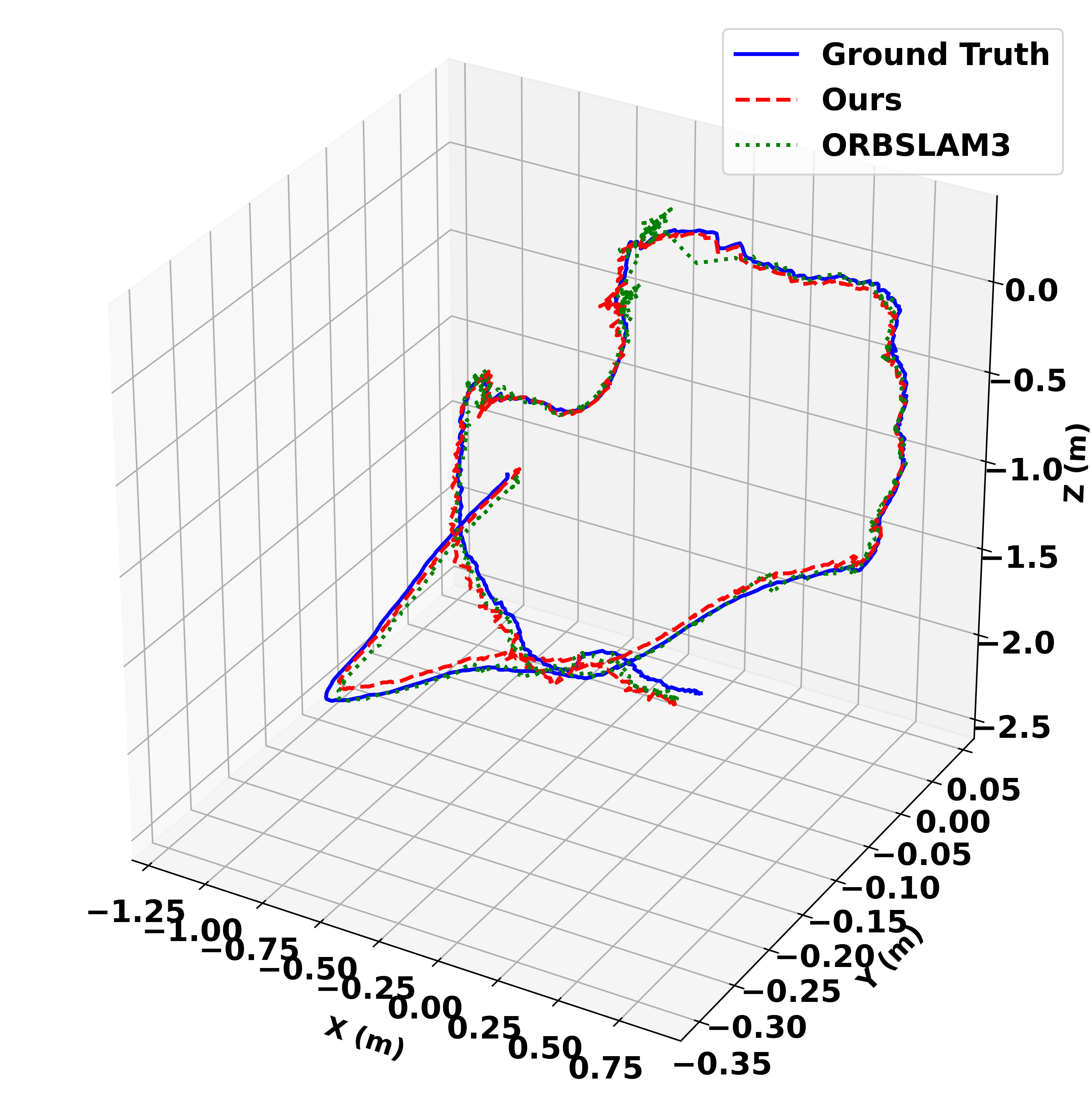}
        \caption{of2}
        \label{fig:subfig5}
    \end{subfigure}
    \hfill
    \begin{subfigure}{0.3\linewidth}
        \includegraphics[width=\linewidth]{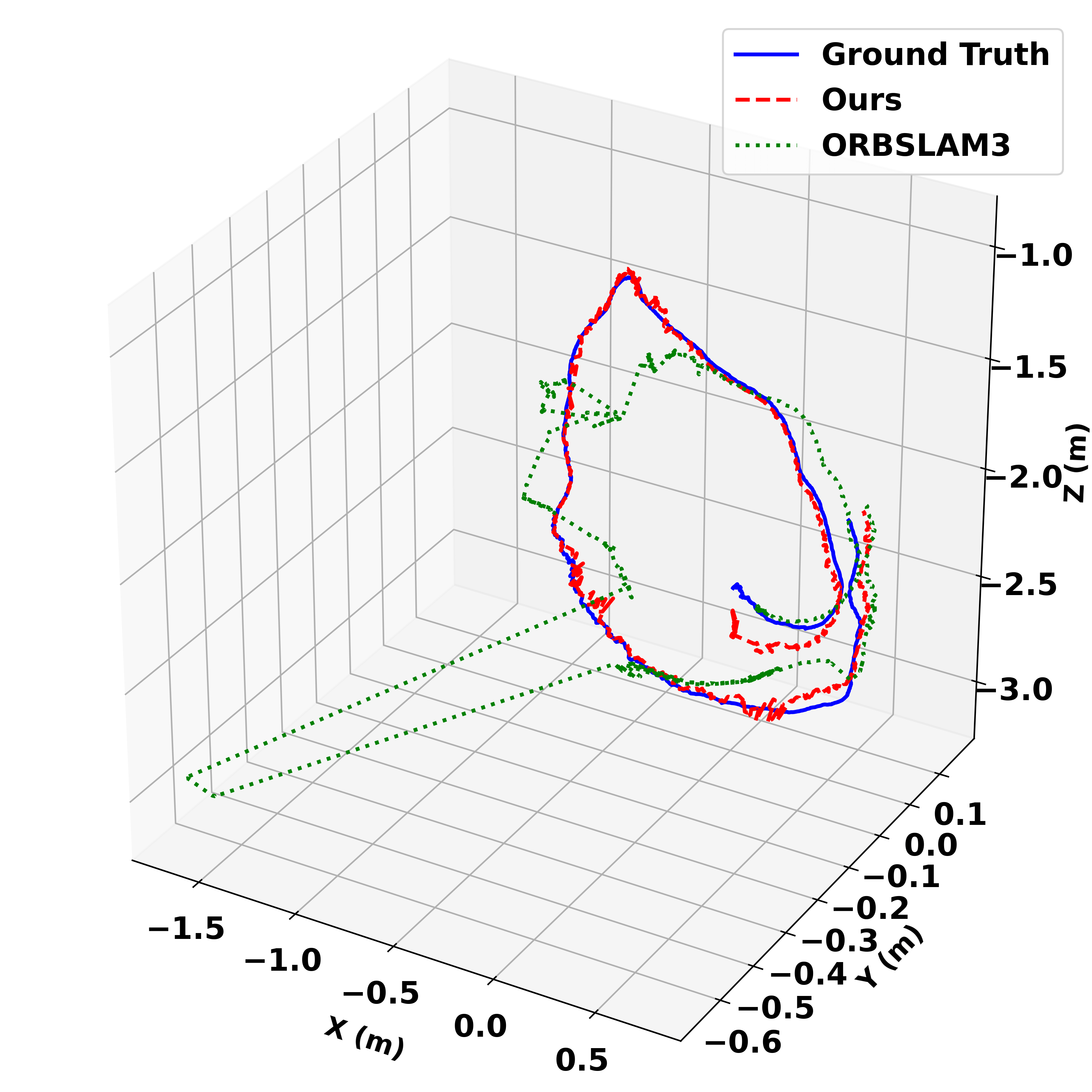}
        \caption{of3}
        \label{fig:subfig6}
    \end{subfigure}

    \caption{3D Trajectory Comparison on ICL-NUIM}
    \label{fig:traj_icl}
\end{figure}

\subsection{Comparison on ICL-NUIM and EuRoC Datasets}
\textbf{Effectiveness of Global Primitives for Monocular SLAM}.    
The ICL-NUIM dataset provides a challenging testbed for monocular SLAM due to its low-contrast and low-texture sequences. As shown in Table~\ref{tab:icl_nuim_results}, our method (denoted as \( -p \) \( -l \) \( -v \) achieves the best results on 4 out of 6 sequences, outperforming state-of-the-art systems such as LSD-SLAM, CNN-SLAM, LPVO, GeoNet, Structure-SLAM, and ORB-SLAM3. This demonstrates the robustness of our approach in handling environments where traditional point-based methods often fail. The integration of line segments and vanishing points, combined with a well-formulated reprojection error, significantly enhances pose estimation accuracy and reliability.  

For further validation, we evaluate our method on the EuRoC dataset, which includes complex and dynamic indoor sequences. As illustrated in Table~\ref{tab:my_label}, our method consistently outperforms other state-of-the-art systems, including PL-SLAM, UV-SLAM, Struct-VIO, PLF-VINS, Structure-PLP-SLAM, and AirVIO. Notably, our system achieves the best results on 4 out of 5 sequences, with an average translational error 12.7\% lower than the second-best system, PL-SLAM. This improvement underscores the effectiveness of integrating vanishing points alongside point and line features, which enhances the system’s ability to exploit structural regularities and improve overall accuracy.
\textbf{Key Advantages of Our Method}.  
The superior performance of our method can be attributed to the following factors:   \textbf{Robust Feature Integration}: By combining point, line, and vanishing point features, our system leverages multiple geometric cues, improving robustness in low-texture and dynamic environments.  \textbf{Well-Formulated Reprojection Error}: Our carefully designed error formulation ensures accurate and stable optimization, particularly for line segments and vanishing points.  \textbf{Structural Regularity Utilization}: The inclusion of vanishing points allows our system to exploit structural regularities in the environment, enhancing pose estimation accuracy in structured indoor scenes.

These results highlight the significant advantages of our approach, demonstrating its potential to advance the state of the art in monocular SLAM systems, particularly in challenging and structured environments.

\begin{table}
\vspace{.2cm}
    \centering
    \caption{Comparison of Translational Error (RMSE) on the ICL-NUIM using monocular camera (unit: m). The best results are highlighted in \textbf{Bold}. }
    \label{tab:icl_nuim_results}
    \resizebox{\linewidth}{!}{
    \begin{tabular}{l|cccccc}
        \toprule
        Monocular & lr-kt1 & lr-kt2 & lr-kt3 & of-kt1 & of-kt2 & of-kt3 \\
        \midrule
        \( -d \) LSD-SLAM~\cite{engel2014lsd} & 0.059 & 0.323 & \( - \) & 0.157 & 0.213 & \( - \) \\
        \( -p \) CNN-SLAM~\cite{tateno2017cnn} & 0.540 & 0.211 & \( - \) & 0.790 & 0.172 & \( - \) \\
        \( -p \) LPVO~\cite{kim2018low} & 0.040 & 0.030 & 0.100 & 0.050 & 0.040 & 0.030\\
        \( -w \) GeoNet~\cite{qi2018geonet} & $\times$ & 0.047 & 0.026 & $\times$ & 0.048 & 0.043 \\
        \( -l \)\( -w \) Structure-SLAM~\cite{li2020structure} & \textbf{0.016} & 0.045 & 0.046 & $\times$ & 0.031 & 0.065 \\
        \( -p \) ORB-SLAM3~\cite{campos2021orb} & $\times$ & 0.037 & \textbf{0.023} & 0.695 & 0.023 & 0.291 \\
        \( -p \)\( -l \)\( -v \) MonoSLAM (Ours) & 0.017 & \textbf{0.022} & 0.055 & \textbf{0.060} & \textbf{0.022} & \textbf{0.038}\\
        \bottomrule
    \end{tabular}}
    
    \vspace{0.5em} 
    \textbf{Note:} \( - \) indicates that the experimental data is not included. $\times$ indicates that the algorithm fails due to lost tracking.
    \( -d \) means that the proposed framework uses direct features, 
    \( -p \) means that the proposed framework uses keypoint feature, 
    \( -l \) means that the proposed framework uses line feature, 
    \( -w \) means that the proposed framework uses the corresponding surface normals, 
    \( -v \) means that the proposed framework uses the vanishing point feature.
\end{table}

\begin{table}[]
    \centering
    \caption{Comparison of Translational Error (RMSE) on the EuRoC dataset (unit: m). The best results are highlighted in \textbf{Bold}.}
    \label{tab:my_label}
    \scriptsize 
    \renewcommand{\arraystretch}{1.2} 
    \resizebox{\linewidth}{!}{
    \begin{tabular}{l|ccc|cccccc}
         \toprule
         \multicolumn{1}{c|}{Methods} & \multicolumn{3}{c|}{Features} & \multicolumn{6}{c}{Sequence} \\
         \cmidrule(lr){2-4} \cmidrule(lr){5-10}
         & \( P \) & \( L \) & \( V \) & MH01 & MH02 & MH03 & MH04 & MH05 & Avg \\ 
         \midrule
         \multicolumn{1}{c|}{AirVIO~\cite{xuairvo}} & $\checkmark$ & $\checkmark$ & $\times$ & 0.074 & 0.060 & 0.114 & 0.167 & 0.125 & 0.108 \\
         \multicolumn{1}{c|}{Structure-PLP-SLAM~\cite{shu2023structure}} & $\checkmark$ & $\checkmark$ & $\times$ & 0.046 & 0.056 & 0.048 & 0.071 & 0.071 & 0.058 \\
         \multicolumn{1}{c|}{PLF-VINS~\cite{lee2021plf}} & $\checkmark$ & $\checkmark$ & $\times$ & 0.143 & 0.178 & 0.221 & 0.240 & 0.260 & 0.208 \\
         \multicolumn{1}{c|}{Struct-VIO~\cite{zou2019structvio}} & $\checkmark$ & $\checkmark$ & $\times$ & 0.119 & 0.100 & 0.283 & 0.275 & 0.256 & 0.207 \\
         \multicolumn{1}{c|}{UV-SLAM~\cite{lim2022uv}} & $\times$ & $\checkmark$ & $\checkmark$ & 0.159 & 0.094 & 0.189 & 0.261 & 0.188 & 0.178 \\
         \multicolumn{1}{c|}{PL-SLAM~\cite{gomez2019pl}} & $\checkmark$ & $\checkmark$ & $\times$ & 0.042 & 0.052 & 0.040 & \textbf{0.064} & 0.070 & 0.053 \\
         \multicolumn{1}{c|}{MonoSLAM (Ours)} & $\checkmark$ & $\checkmark$ & $\checkmark$ & \textbf{0.041} & \textbf{0.041} & \textbf{0.036} & 0.066 & \textbf{0.054} & \textbf{0.047} \\
         \bottomrule
    \end{tabular}}
    
    \vspace{0.5em} 
    \textbf{Note:} \( P \) denotes the keypoint feature, \( L \) denotes the line feature, and \( V \) denotes the vanishing point feature.
\end{table}

The Figure~\ref{fig:traj_icl} provides a comprehensive comparison of the tracking performance between our monocular SLAM system and ORBSLAM3 on the ICL-NUIM dataset. As illustrated, our system exhibits superior performance, achieving significantly higher tracking accuracy across various scenarios. This improvement can be attributed to the enhanced robustness of our approach, particularly in challenging environments. In contrast, ORBSLAM3, which relies exclusively on point features for tracking, is susceptible to tracking failures and scale drift, especially in low-texture regions of the dataset. The lack of sufficient point features in such environments severely compromises its stability and reliability, highlighting the limitations of a point-feature-only framework. Our system, on the other hand, leverages additional geometric constraints and feature types, ensuring consistent performance even in texture-poor scenarios. This comparative analysis underscores the effectiveness of our design in addressing the limitations of traditional point-feature-based SLAM systems.

\begin{table}[]
\vspace{.2cm}
    \centering
    \caption{Comparison of Translational Error (RMSE) on the EuRoC dataset (unit: m). The best results are highlighted in \textbf{Bold}.}
    \label{tab:ablation}
    \resizebox{\linewidth}{!}{
    \begin{tabular}{l|ccc|cccccc}
         \toprule
         \multicolumn{1}{c|}{Ablation studies} & \multicolumn{3}{c|}{Features} & \multicolumn{6}{c}{Sequence} \\
         \cmidrule(lr){2-4} \cmidrule(lr){5-10}
         & \( P \) & \( L \) & \( V \) & MH01 & MH02 & MH03 & MH04 & MH05 & Avg \\ 
         \midrule
        \multicolumn{1}{c|}{MonoSLAM (\( LP \))} & $\checkmark$ & $\checkmark$ & $\times$ & 0.052 & \textbf{0.037} & 0.047 & 0.109 & 0.095 & 0.067 \\
         \multicolumn{1}{c|}{MonoSLAM (\( GP \))} & $\checkmark$ & $\checkmark$ & $\checkmark$ & \textbf{0.041} & 0.041 & \textbf{0.036} & \textbf{0.066} & \textbf{0.054} & \textbf{0.047} \\
         \bottomrule
    \end{tabular}}

    \vspace{0.5em} 
    \textbf{Note:} MonoSLAM (\( LP \)) denotes our system using optimization based on Local Primitives, MonoSLAM (\( GP \)) denotes our system using optimization based on both Local Primitives and Global Primitives.
\end{table}

\subsection{Ablation Studies}
\label{sec:ablation_studies}

In this section, we perform ablation studies to quantify the impact of Global Primitives factor graph optimization on tracking accuracy of MonoSLAM. Through systematic comparisons, we evaluate our system performance using optimization based solely on LP versus the combined optimization of LP and GP across the EuRoC dataset. The results, summarized in Table~\ref{tab:ablation}, detail the RMSE for each sequence. The integration of Global Primitives factor graph optimization demonstrates consistent improvement in translational accuracy across most sequences. Specifically, the errors for sequences MH01, MH03, MH04 and MH05 are reduced by 21.2\%, 23.4\%, 39.4\%, and 43.2\%, respectively. However, sequence MH02 exhibits an unexpected increase of 9.75\%. This anomaly can be primarily attributed to the sequence’s unique characteristics, particularly the absence of strong vanishing direction features, which limits the effectiveness of Global Primitives factor graph  optimization in this specific case.

The overall improvement in tracking accuracy can be attributed to the ability of Global Primitives factor graph optimization to jointly optimize multiple frames by leveraging parallel vanishing directions in the world coordinate system. This cross-frame constrained optimization enhances the accuracy of pose estimation by providing additional geometric constraints, which are particularly beneficial in environments with prominent linear features.

In summary, the ablation studies demonstrate that Global Primitives factor graph optimization significantly enhances the performance of our MonoSLAM system, as evidenced by the reduction in RMSE across most sequences. This improvement underscores the importance of incorporating geometric constraints from vanishing directions in visual SLAM systems, particularly in structured environments where such features are prevalent.

\section{Discussion and feature work}
We propose a monocular SLAM system based on point, line, and vanishing  point features, which utilizes global primitives to associate  multi-frame non-overlapping images and incorporates a novel factor graph optimization. Our system achieves state-of-the-art performance. We have demonstrated that leveraging vanishing points extracted from a single RGB image can significantly improve pose estimation accuracy without  relying on environmental assumptions. Compared to other advanced  real-time monocular SLAM methods, our approach struggles to maintain  high stability under high dynamic motions (e.g., rapid acceleration or  deceleration) in the absence of IMU data. In the future, integrating IMU information could be explored to further refine camera pose estimation.